\documentclass{ecai2004}

\usepackage{times}
\usepackage{graphicx}
\usepackage{latexsym}

\usepackage{epsf,psfig}

\usepackage{algorithmic,algorithm} 
\usepackage{amsmath}

\newcommand{\commentaire}[1]{ } 
\newcommand{\definition}[1]{\textbf{Definition:} #1}

\title{How to use the Scuba Diving metaphor to solve problem with neutrality ?}

\author{Collard Philippe \and Verel S\'ebastien \and Clergue Manuel
\institute{Laboratoire I3S, Universit\'{e} de Nice-Sophia Antipolis, France,
Email: \{pc, verel, clerguem\}@i3s.unice.fr}}

\begin{document}

\maketitle

\bibliographystyle{ecai2004}

\begin{abstract}
We proposed a new search heuristic using the \textit{scuba diving} metaphor. This approach is based on the concept of evolvability and tends to exploit neutrality which exists in many real-world problems. Despite the fact that natural evolution does not directly select for evolvability, the basic idea behind the \textit{scuba search} heuristic is to explicitly push evolvability to increase. A comparative study of the scuba algorithm and standard local search heuristics has shown the advantage and the limitation of the scuba search. In order to tune neutrality, we use the $NKq$ fitness landscapes and a family of travelling salesman problems (TSP) where cities are randomly placed on a lattice and where travel distance between cities is computed with the Manhattan metric. In this last problem the amount of neutrality varies with the city concentration on the grid ; assuming the concentration below one, this TSP reasonably remains a NP-hard problem. 
\end{abstract}

\section{Introduction}
In this paper we propose an heuristic called \textit{Scuba Search} that allows us to exploit the neutrality that is present in many real-world problems. This section presents the interplay between neutrality in search space and metaheuristics. Section 2 describes the {\it Scuba Search} heuristic in details. In order to illustrate efficiency and limit of this heuristic, we use the $NKq$ fitness landscapes and a travelling salesman problem (TSP) on diluted lattices as a model of neutral search space. These two problems are presented in section 3. Experiment results are given in section 4 where comparisons are made with two hill climbing heuristics. In section 5, we point out advantage and shortcoming of the approach; finally, we summarize our contribution and present plans for a future work.

\subsection{Neutrality}

The metaphor of an 'adaptative landscape' introduced by S.~Wright~\cite{wright:rmicse} has dominated the view of adaptive evolution: an 
uphill walk of a population on a mountainous fitness landscape in which 
it can get stuck on suboptimal peaks.
Results from molecular evolution has changed this picture: Kimura's model~\cite{KIM:83} assumes that the overwhelming majority of mutations are 
either effectively neutral or lethal and in the latter case purged by 
negative selection. This assumption is called the neutral hypothesis. 
Under this hypothesis, the dynamics of populations evolving on such 
neutral landscapes are different from those on adaptive landscapes: they 
are characterized by long periods of fitness stasis (population is situated on a 'neutral network') punctuated by shorter periods of innovation with rapid fitness 
increase.
In the field of evolutionary computation, neutrality plays an important role in real-world problems: in design of digital circuits 
\cite{vassilev00advantages}, in evolutionary robotics \cite{jakobi95noise}. In those problems, neutrality is implicitly embedded in the genotype to phenotype mapping.

\subsection{Evolvability}

\textit{Evolvability} is defined by Altenberg~\cite{WA-AL} as \textquotedblleft the ability of random variations to sometimes produce improvement\textquotedblright . This concept refers to the efficiency of evolutionary search; it is based upon the work by Altenberg~\cite{ALT:94}: \textquotedblleft the ability of an operator/representation scheme to produce offspring that are fitter than their parents\textquotedblright . As enlighten by Turney~\cite{turney99increasing} the concept of evolvability is difficult to define. As he puts it: \textquotedblleft if $s$ and $s{'}$ are equally fit, $s$ is more \textit{evolvable} than $s{'}$ if the fittest offspring of $s$ is more likely to be fitter than the fittest offspring of $s{'}$\textquotedblright . Following this idea we define evolvability as a function (see section \ref{def_evol}). 




\section{Scuba Search}

The Scuba Search, heuristic introduced in this section, exploits neutrality by combining local search heuristic with navigating the neutral neighborhood of states.

\subsection{The Scuba Diving Metaphor}
Keeping the landscape as a model, let us imagine this landscape with peaks (local optima) and lakes (neutral networks).  Thus, the landscape is bathed in
an uneven sea; areas under water represent non-viable solutions. So 
there are paths from one peak to another one for a swimmer. The key, of
course, remains to locate an attractor which represents the system's maximum
fitness. In this context, the problem is to find out how to cross a lake without global
information. We use the scuba diving metaphor as a guide to present
principles of the so-called \textit{scuba search} ($SS$). This heuristic is a
way to deal with the problem of crossing in between peaks. then we avoid to be
trapped in the vicinity of local optima. The problem is to get to know what drives the swimmer
from one edge of the lake to the opposite one? Up to the classic view a swimmer drifts at the surface of a lake. The new metaphor is a scuba
diving seeing the \textit{world above the water surface}. We propose a new heuristic to cross a neutral net getting information above-the-surface (ie. from fitter points in the neighborhood).
\subsection{Scuba Search Algorithm}

Despite the fact that natural evolution does not directly select for evolvability, there is a dynamic pushing evolvability to increase~\cite{turney99increasing}. The basic idea behind the $SS$ heuristic is to explicitly push evolvability to increase. Before presenting this search algorithm, we need to introduce a new type of local optima, the {\it local-neutral optima}. Indeed with $SS$ heuristic, local-neutral optima will allow transition from neutral to adaptive evolution. So evolvability will be locally optimized. Given a search space ${\cal S}$ and a fitness function $f$ defined on ${\cal S}$, some more precise definitions follow.

\definition{A {\it neighborhood structure} is a function ${\cal V} : {\cal S} \rightarrow 2^{\cal S}$ that assigns to every $s \in {\cal S}$ a set of neighbors ${\cal V}(s)$ such that $s \in {\cal V}(s)$.}

\label{def_evol}
\definition{The {\it evolvability} of a solution $s$ is the function $evol$ that assigns to every $s \in {\cal S}$ the maximum fitness from the neighborhood ${\cal V}(s)$: $\forall s \in {\cal S}$, $evol(s) = max \lbrace f(s^{'}) ~|~ s^{'} \in  {\cal V}(s) \rbrace$.}
\commentaire{Definition : A {\it local maximum} is a solution $s_{opt}$ such that $ \forall s^{'} \in {\cal V}(s_{opt})$, $ f(s^{'}) \leq f(s_{opt})$. We call $s_{opt}$ a {\it strict local maximum} if $ \forall s^{'} \in {\cal V}(s_{opt})$, $ f(s^{'}) < f(s_{opt})$\\
}

\definition{For every fitness function $g$, neighborhood structure ${\cal W}$ and genotype $s$, the predicate $isLocal$ is defined as:\\
$isLocal(s, g, {\cal W}) = (\forall s^{'} \in {\cal W}(s), g(s^{'}) \leq g(s) )$.}

\definition{For every $s \in {\cal S}$, the {\it neutral set} of $s$ is the set ${\cal N} (s) = \lbrace s^{'} \in {\cal S}~|~ f(s^{'}) = f(s) \rbrace$, and the {\it neutral neighborhood} of $s$ is the set ${\cal V}n(s) = {\cal V}(s) \cap {\cal N}(s)$.}

\definition{For every $s \in {\cal S}$, the {\it neutral degree} of $s$, noted $Degn(s)$, is the number of neutral neighbors of $s$, $Degn(s) = \#{\cal V}n(s) - 1$.}

\definition{A solution $s$ is a \textit{local maximum} iff $isLocal(s, f, {\cal V})$.}

\definition{A solution $s$ is a \textit{local-neutral maximum} iff $isLocal(s, evol, {\cal V}n)$.}

Scuba Search use two dynamics one after another (see algo.\ref{algoScuba}).
The first one corresponds to a neutral path. At each step the scuba diving remains under the water surface driven by the hands-down fitnesses; that is fitter fitness values reachable from one neutral neighbor. At that time the \textit{flatCount} counter is incremented. When the diving reaches a local-neutral optimum, \textit{i.e.} when all the fitnesses reachable from one neutral neighbor are selectively neutral or disadvantageous, the neutral path stops and the diving starts up the \textit{Invasion-of-the-Land}. Then the \textit{gateCount} counter increases. This process goes along, switching between \textit{Conquest-of-the-Waters} and \textit{Invasion-of-the-Land}, until a local optimum is reached.
\begin{algorithm}
\caption{Scuba Search}
\label{algoScuba}
\begin{algorithmic}
\STATE flatCount $\leftarrow$ 0, gateCount $\leftarrow$ 0
\STATE Choose initial solution $s \in \cal S$
\REPEAT

	\WHILE{not $isLocal(s, evol, {\cal V}n)$}
		\STATE $M = max \lbrace evol(s^{'})~|~s^{'} \in {\cal V}n(s)-\{s\} \rbrace$
		\IF{$evol(s) < M$}
			\STATE choose $s^{'} \in {\cal V}n(s)$ such that $evol(s^{'}) = M$
			\STATE $s \leftarrow s^{'}$, flatCount $\leftarrow$ flatCount +1
		\ENDIF
	\ENDWHILE
	\STATE choose $s^{'} \in {\cal V}(s) - {\cal V}n(s)$ such that $f(s^{'}) = evol(s)$
	\STATE $s \leftarrow s^{'}$, gateCount $\leftarrow$ gateCount +1
\UNTIL{$isLocal(s, f, {\cal V})$}
\end{algorithmic}
\end{algorithm}

\section{Models of Neutral Seach Space}
In order to study the Scuba Search heuristic we have to use landscapes with a tunable degree of neutrality.

\subsection{The NKq fitness Landscape}
The $NKq$ fitness landscapes family proposed by Newman {\it et al}.~\cite{newman98effect} has properties of systems undergoing neutral selection such as RNA sequence-structure maps. 
It is a generalization of the $NK$-landscapes proposed by Kauffman~\cite{KAU:93} where parameter $K$ tunes the ruggedness and parameter  $q$ tunes the degree of neutrality.


\subsubsection{Definition and properties}
The fitness function of a $NKq$-landscape~\cite{newman98effect} is a function $f: \lbrace 0, 1 \rbrace^{N} \rightarrow [0,1]$ defined on binary strings with $N$ loci.
Each locus $i$ represents a gene with two possible alleles, $0$ or $1$. 
An 'atom' with fixed epistasis level is represented by a fitness components $f_i: \lbrace 0, 1
\rbrace^{K+1} \rightarrow [0,q-1]$ associated to each locus $i$. It depends
on the allele at locus $i$ and also on the alleles at $K$ other epistatic loci
($K$ must fall between $0$ and $N - 1$). The fitness $f(x)$ of $x \in \lbrace 0, 1 \rbrace^{N}$
is the average of the values of the $N$ fitness components $f_i$:\label{defNK}

$$ f(x) = \frac{1}{N (q-1)} \sum_{i=1}^{N} f_i(x_i; x_{i_1}, \ldots, x_{i_K})
$$ where $\lbrace i_1, \ldots, i_{K} \rbrace \subset \lbrace 1, \ldots, i -
1, i + 1, \ldots, N \rbrace$. Many ways have been proposed to choose the $K$ other
loci from $N$ loci in the genotype. Two possibilities are mainly used: adjacent and
random neighborhoods. With an adjacent neighborhood, the $K$ genes
nearest to the locus $i$ are chosen (the genotype is taken to have periodic boundaries).
With a random neighborhood, the $K$ genes are chosen randomly on the genotype.
Each fitness component $f_i$ is specified by extension, ie an integer number $y_{i,(x_i; x_{i_1}, \ldots, x_{i_K})}$ from $[0, q - 1]$ 
is associated with each element $(x_i; x_{i_1}, \ldots, x_{i_K})$ from $\lbrace 0, 1 \rbrace^{K+1}$.
Those numbers are uniformly distributed in the interval $[0, q - 1]$.
The parameters of $NKq$-landscape tune ruggedness and neutrality of the landscape \cite{nic-comparison}. The number of local optima is linked to the parameter $K$. The largest number is obtained when $K$ takes its maximum value $N-1$. The neutral degree (see tab. \ref{tab_Degn}) decreases as $q$ or $K$ increases. The maximal degree of neutrality appears when $q$ takes value $2$.

\begin{table}
\caption{Average neutral degree on $NKq$-landscapes with $N=64$ performs on $50000$ genotypes}
\begin{center}
\begin{tabular}{c|cccccc}
\hline
 & \multicolumn{6}{|c}{K} \\ 
\cline{2-7}
q & 0 & 2 & 4 & 8 & 12 & 16 \\
\hline
2 & 35.00 & 21.33 & 16.56 & 12.39 & 10.09 & 8.86 \\
3 & 21.00 & 13.29 & 10.43 & 7.65 & 6.21 & 5.43 \\
4 & 12.00 & 6.71 & 4.30 & 2.45 & 1.66 & 1.24 \\
100 & 1.00 & 0.32 & 0.08 & 0.00 & 0.00 & 0.00 \\
\hline
\end{tabular}	
\end{center}
\label{tab_Degn}
\end{table}

\subsubsection{Parameters setting}

All the heuristics used in our experiments are applied to a same instance of $NKq$ fitness landscapes\footnote{With random neighborhood} with $N=64$. The neighborhood is the classical one-bit mutation neighborhood: ${\cal V}(s) = \lbrace s^{'}~|~ Hamming(s^{'}, s) \leq 1 \rbrace$. For each triplet of parameters $N$, $K$ and $q$, $10^3$ runs were performed.

\subsection{The Travelling Salesman Problem on randomly diluted lattices}
The family of TSP proposed by Chakrabarti~\cite{CHA:00} is an academic benchmark that allows to test our ideas. These problems do not reflect the true reality but is a first step towards more real-life benchmarks. We use these problems to incorporate a tunable level of neutrality into TSP search spaces.

\subsubsection{Definition and properties}
The travelling salesman problem is a well-known combinatorial optimization problem: given a finite number $N$ of cities along with the cost of travel between each pair of them, find the cheapest way of visiting all the cities and returning to your starting point. In this paper we use a TSP defined on randomly dilute lattices. The $N$ cities randomly occupy lattice sites of a two-dimentional square lattice ($L \times L$). We use the \textit{Manhattan} metric to determine the distance between two cities. The lattice \textit{occupation concentration} (i.e. the fraction of sites occupied) is $\frac{N}{{^{\mathop L\nolimits^2 } }}$. As the concentration is related to the \textit{neutral degree}, we note $TSPn$ such a problem with concentration $n$.
For $n=1$, the problem is trivial as it can be reduced to the \textit{one-dimensional} TSP. As $n$ decreases from unity the problem becomes nontrivial: the discreteness of the distance of the path connecting two cities and the angle which the path makes with the Cartesian axes, tend to disappear. Finally, as $n \to 0$, the problem can be reduced to the standard \textit{two-dimensional} TSP. As Chakrabarti~\cite{CHA:00} stated: \textquotedblleft it is clear that the problem crosses from triviality (for $n=1$) to NP-hard problem at a certain value of $n$. We did not find any irregularity $\left| {...} \right|$ at any $n$. The crossover from triviality to NP-hard problem presumably occurs at $n=1$.\textquotedblright

\begin{figure}
\psfig{figure=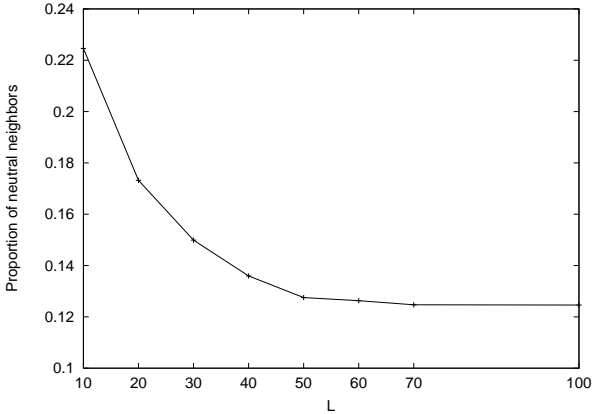,width=230pt,height=160pt}
\caption{Average proportion of neutral neighbors on $TSPn$ as function of L, for $N=64$ (values are computing from $50000$ random solutions)}
\label{fig_Degn}
\end{figure}

The idea is to discretize the possible distances, through only allowing each distance to take one of $D$ distances. Varying this terrace parameter $D$ from an infinite value (corresponding to the standard TSP), down to the minimal value of $1$ thus decreases the number of possible distances, so increasing the fraction of equal fitness neutral solutions. So, parameter $n=\frac{N}{L^2}$ of the $TSPn$ tunes both the concentration and the neutral degree (see fig. \ref{fig_Degn}). In the remaining of this paper we consider $TSPn$ problems where $n$ stands in the range $\left[ {0,1} \right]$.

\subsubsection{Parameters setting}
All the heuristics used in our experiments are applied to a same instance of $TSPn$. The search space ${\cal S}$ is the set of permutations of $\lbrace 1, \ldots, N \rbrace$. The neighborhood is induced by the classical 2-$opt$ mutation operator: ${\cal V}(s) = \lbrace s^{'}~|~ s^{'} = \text{2-}opt(s) \rbrace$. The size of neighborhood is then $\frac{N(N-3)}{2}$. For each value of $L$, $500$ runs were performed.

\section{Experiment Results}

\subsection{Algorithm of Comparison}
\label{sec_algo}
Two {\it Hill Climbing} algorithms are used for comparison.

\subsubsection{Hill Climbing}
The simplest type of local search is known as \textit{Hill Climbing} ($HC$) when trying to maximize a solution. $HC$ is very good at exploiting the neighborhood; it always takes what looks best at that time. But this approach has some problems. The solution found depends on the initial solution. Most of the time, the found solution is only a local optima.
We start off with a probably suboptimal solution. We then look in the neighborhood of that solution to see if there is something better. If so, we adopt this improved solution as our current best choice and repeat. If not, we stop assuming that the current solution is good enough (local optimum).
\begin{algorithm}
\caption{Hill Climbing}
\label{algoHC}
\begin{algorithmic}
\STATE step $\leftarrow$ 0
\STATE Choose initial solution $s \in \cal S$
\REPEAT
	\STATE choose $s^{'} \in {\cal V}(s)$ such that $f(s^{'}) = evol(s)$
	\STATE $s \leftarrow s^{'}$, step $\leftarrow$ step + 1
\UNTIL{$isLocal(s, f, {\cal V})$}
\end{algorithmic}
\end{algorithm}

\subsubsection{Hill Climbing Two Steps}

Hill Climber can be extended in many ways. {\it Hill Climber two Step} ($HC2$) exploits a larger neighborhood of stage 2. The algorithm is nearly the same as $HC$. $HC2$ looks in the extended neighborhood of stage two of the current solution to see if there is something better. If so, $HC2$ adopts the solution in the neighborhood of stage one which can reach a best solution in the extended neighborhood. If not, $HC2$ stop assuming the current solution is good enough.
So, $HC2$ can avoid more local optimum than $HC$. 
Before presenting the algorithm \ref{algoHCDeuxPas} we must introduce the following definitions:

\definition{The {\it extended neighborhood structure\footnote{Let's note that ${\cal V}(s) \subset {\cal V}^2(s)$}} from ${\cal V}$ is the function ${\cal V}^2(s) = \cup_{s_1 \in {\cal V}(s)} {\cal V}(s_1)$}

\definition{$evol^2$ is the function that assigns to every $s \in {\cal S}$ the maximum fitness from the extended neighborhood ${\cal V}^2(s)$. $\forall s \in {\cal S}$, $evol^2(s) = max \lbrace f(s^{'}) | s^{'} \in {\cal V}^2(s) \rbrace$}

\begin{algorithm}
\caption{Hill Climbing (Two Steps)}
\label{algoHCDeuxPas}
\begin{algorithmic}
\STATE step $\leftarrow$ 0
\STATE Choose initial solution $s \in \cal S$
\REPEAT
\IF{$evol(s)=evol^2(s)$}
	\STATE choose $s^{'} \in {\cal V}(s)$ such that $f(s^{'}) = evol^2(s)$
\ELSE
	\STATE choose $s^{'} \in {\cal V}(s)$ such that $evol(s^{'}) = evol^2(s)$
\ENDIF
\STATE $s \leftarrow s^{'}$, step $\leftarrow$ step + 1
\UNTIL{$isLocal(s, f, {\cal V}^2)$}
\end{algorithmic}
\end{algorithm}

\commentaire{
    Gerhard Reinelt.
    The Travelling Salesman. Computational Solutions for TSP Applications, volume 840 of Lecture Notes in Computer Science.
    Springer-Verlag, Berlin Heidelberg New York, 1994. 
}

\subsection{Performances}
In this section we present the average fitness found using each heuristic on both $NK$ and $TSP$ problems.

\subsubsection{NKq Landscapes}
\begin{figure*}[!tb] 
\begin{center}
\begin{tabular}{cc} 
\psfig{figure=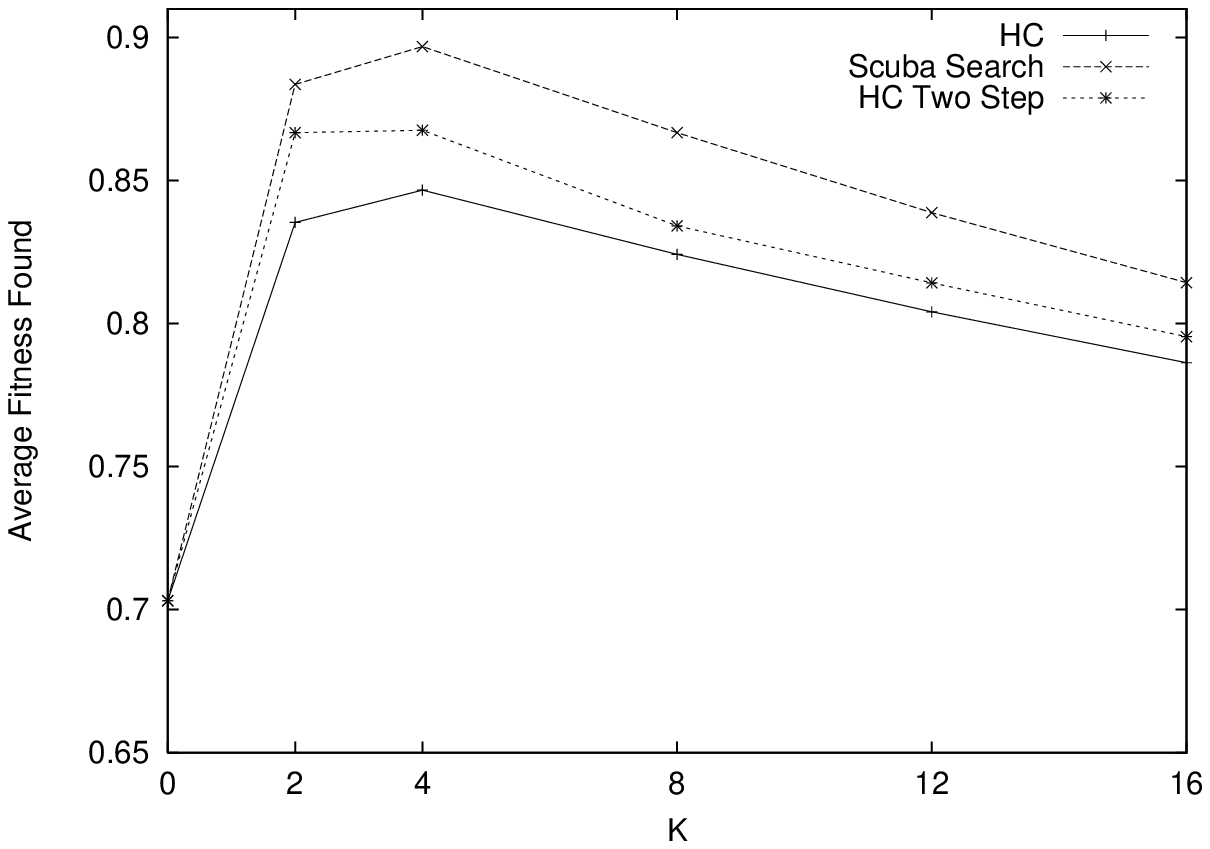,width=230pt,height=160pt} 
&
\psfig{figure=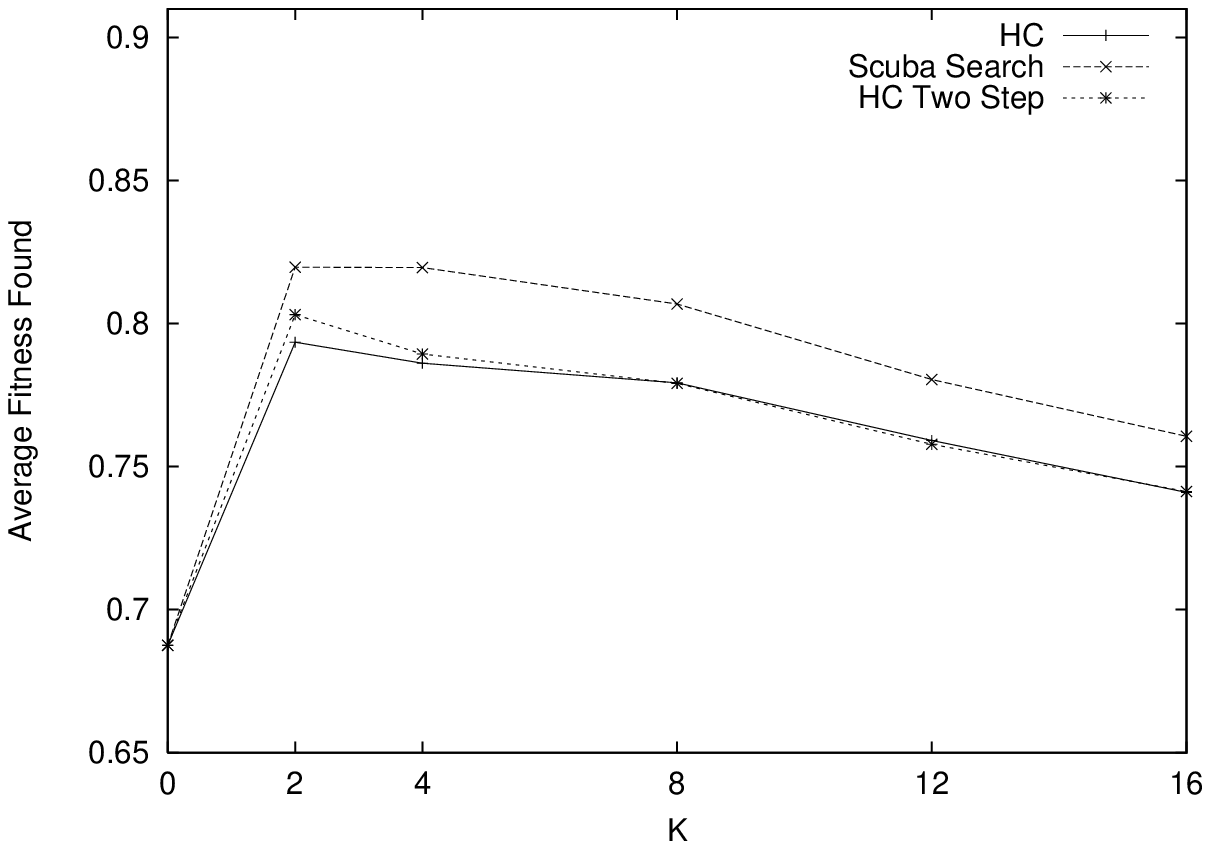,width=230pt,height=160pt}
\\
(a) &
(b) \\
\psfig{figure=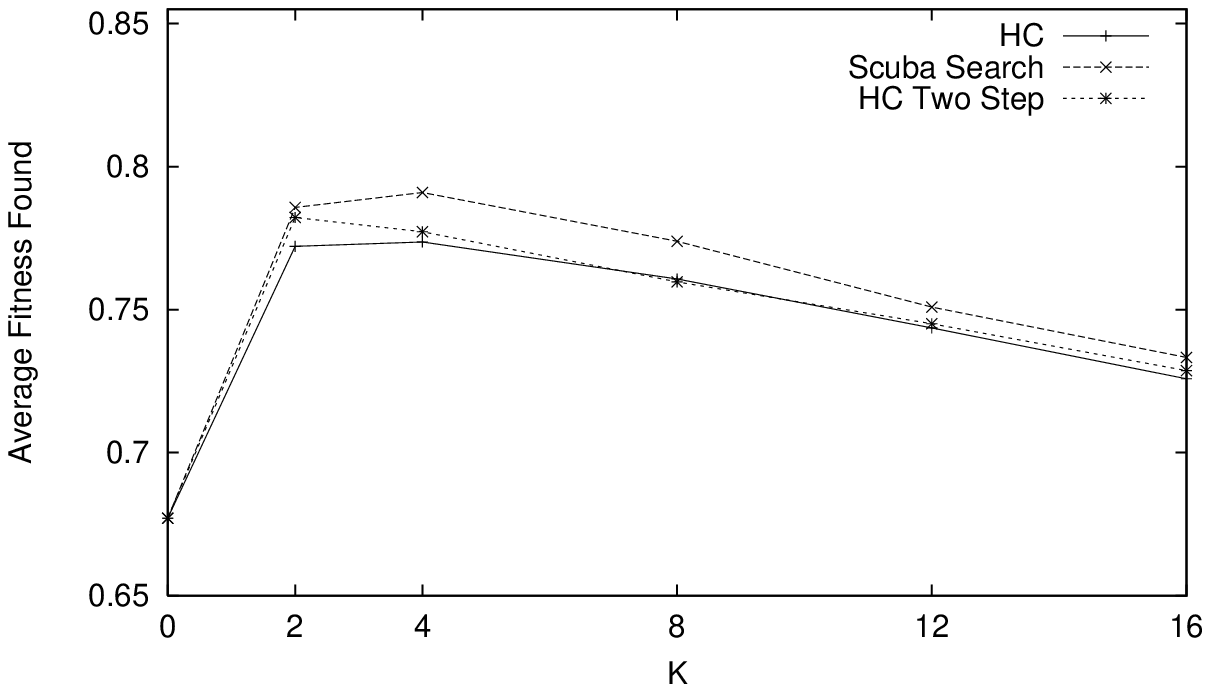,width=230pt,height=130pt}
&
\psfig{figure=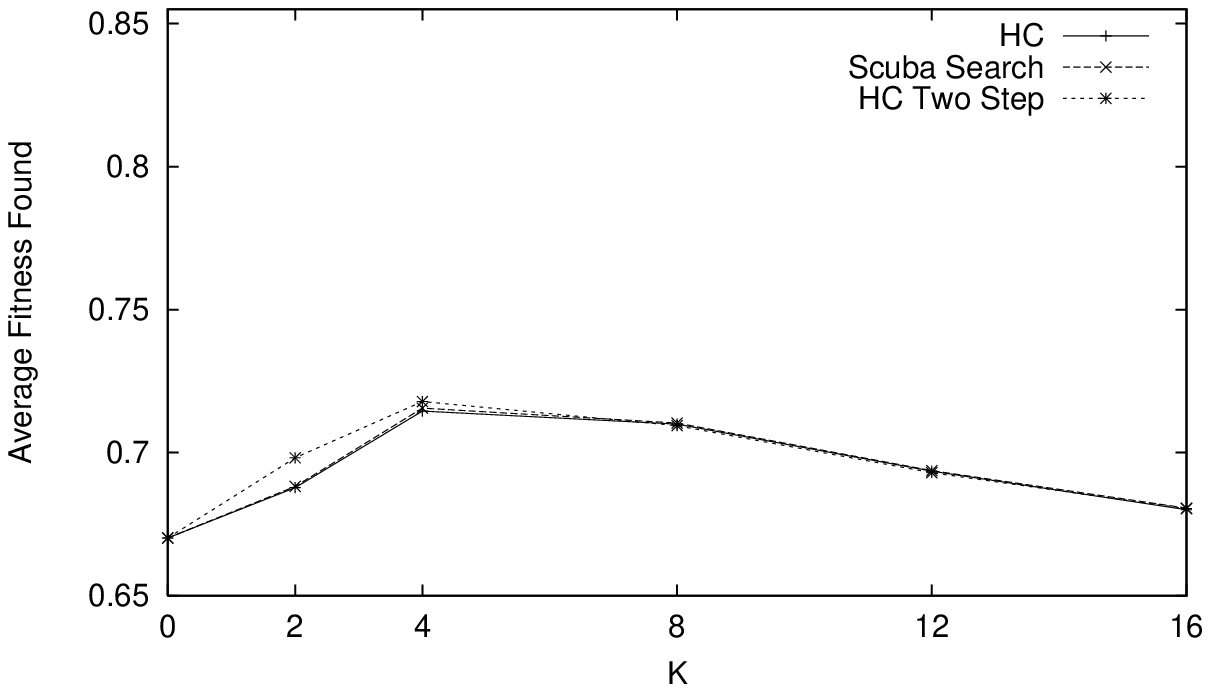,width=230pt,height=130pt} \\
(c) &
(d) \\
\end{tabular}
\end{center}
\caption{Average fitness found on $NKq$-landscapes as function of $K$, for $N=64$ and $q=2$ (a), $q=3$ (b), $q=4$ (c), $q=100$ (d)}
\label{fig_fit_q}
\end{figure*}

Figure \ref{fig_fit_q} shows the average fitness found respectively by each of the three heuristics as a function of the epistatic parameter $K$ for different values of the neutral parameter $q$. In the presence of neutrality, according to the average fitness, \textit{Scuba Search} outperforms \textit{Hill Climbing} and \textit{Hill Climbing two steps}. Let us note that with high neutrality ($q=2$ and $q=3$), the difference is still more significant. Without neutrality ($q=100$) all the heuristics are nearly equivalent. The Scuba Search have on average better fitness value for $q=2$ and $q=3$ than hill climbing heuristics. This heuristic benefits in $NKq$ from the neutral paths to reach the highest peaks.\\

\subsubsection{TSPn Problems}

Table \ref{tab_Perf} shows the fitness performances of heuristics on $TSPn$ landscapes. The average and the best fitness found by $SS$ are always above the ones for $HC$. As for $NKq$ landscapes, the difference is more important when neutrality is more significant ($L=10$ and $L=20$). Performances of $SS$ are a little better for $L=10$ and $L=20$ and a little less for $L=30$ and $L=100$. Let us also note that standart deviation is still smaller for $SS$. 

\begin{table}
\begin{center}
{\caption{Average and standart deviation of fitness found on $TSPn$ ($N=64$) performed on $500$ independants runs. Best fitness found is putted in brackets} \label{tab_Perf}}
\begin{tabular}{c|c c c c}
\hline
heurist & \multicolumn{4}{|c}{L} \\
\cline{2-5}
& 10 & 20 & 30 & 100 \\
\hline
$HC$ & $ 101_{5}(90) $ & $ 193_{10}(164)$ & $ 293_{13}(256) $ & $ 872_{44}(770) $ \\
$SS$ & $ 93_{4}(84) $ & $ 180_{8}(162) $ & $ 281_{12}(254) $ & $ 857_{41}(764) $ \\
$HC2$ & $ 95_{8}(86) $ & $ 184_{15}(162) $ & $ 282_{18}(252) $ & $ 854_{61}(764) $ \\
\hline
\end{tabular}    
\end{center}

\end{table}


\subsection{Evaluation cost}

\subsubsection{NKq Landscapes}
Table \ref{tab_eval_NKq} shows the number of evaluations for the different heuristics. For all the heuristics, the number of evaluations decreases with $K$. The evaluation cost decreases as ruggedness increases. For $HC$ and $HC2$, the evaluation cost increases with $q$. For $HC$ and $HC2$, more neutral the landscape is, smaller the evaluation cost. Conversely, for $SS$ the cost decreases with $q$. 
At each step the number of evaluations is $N$ for $HC$ and $\frac{N(N-1)}{2}$ for $HC2$. So, the cost depends on the length of adaptive walk of $HC$ and $HC2$ only. The evaluation cost of $HC$ and $HC2$ is low when local optima are nearby (i.e. in rugged landscapes).
For $SS$, at each step, the number of evaluations is $(1 + Degn(s))N$ which decreases with neutrality. So, the number of evaluations depends both on the number of steps in $SS$ and on the neutral degree. The evaluation cost of $SS$ is high in neutral landscape.
\begin{table}
\caption{Average number of evaluations on $NKq$-landscape with $N=64$}

\begin{tabular}{l c|cccccc}
\cline{3-8}
 & & \multicolumn{6}{c}{K} \\ 
\cline{3-8}
& q & 0 & 2 & 4 & 8 & 12 & 16 \\
\hline
$HC$      &   & 991 & 961 & 807 & 613 & 491 & 424 \\
$SS$      & 2 & 35769 & 23565 & 15013 & 8394 & 5416 & 3962 \\
$HC2$ &   & 29161 & 35427 & 28038 & 19192 & 15140 & 12374 \\
\hline
$HC$      &   & 1443 & 1159 & 932 & 694 & 546 & 453 \\
$SS$      & 3 & 31689 & 17129 & 10662 & 6099 & 3973 & 2799 \\
$HC2$ &   & 42962 & 37957 & 29943 & 20486 & 15343 & 12797 \\
\hline
$HC$      &   & 1711 & 1317 & 1079 & 761 & 614 & 500 \\
$SS$      & 4 & 22293 & 9342 & 5153 & 2601 & 1581 & 1095 \\
$HC2$ &   & 52416 & 44218 & 34001 & 22381 & 18404 & 14986 \\
\hline
$HC$      &     & 2102 & 1493 & 1178 & 832 & 635 & 517 \\  
$SS$      & 100 & 4175 & 1804 & 1352 & 874 & 653 & 526 \\  
$HC2$ &     & 63558 & 52194 & 37054 & 24327 & 18260 & 15271 \\ 
\hline
\end{tabular}	
\label{tab_eval_NKq}
\end{table}

\subsubsection{TSPn Problems}

Table \ref{tab_res} shows the number of evaluations on $TSPn$. Scuba Search uses a larger number of evaluations than $HC$ (nearly $200$ times on average) and smaller than $HC2$ (nearly $12$ times on average). As expected, for $SS$ the evaluation cost decreases with $L$ and so the neutrality of landscapes; whereas it increase for $HC$ and $HC2$. Landscape seems more rugged when $L$ is larger.

\begin{table}
\begin{center}
{\caption{Average number of evaluations (x $10^6$) on the family of $TSPn$ problems with $N=64$}\label{tab_res}}
\begin{tabular}{c | c c c c}
\hline
& \multicolumn{4}{|c}{L} \\
\cline{2-5}
& 10 & 20 & 30 & 100 \\
\hline
$HC$      & $0.0871$   & $0.101$     & $0.105$     & $0.117$ \\
$SS$      & $25.3$     & $20.2$      & $16.6$      & $13.0$  \\
$HC2$     & $183.7$    & $204.0$     & $211.4$     & $230.5$ \\
\commentaire{
$HC$      & 87114      & 100973      & 104970      & 117241 \\
$SS$      & 2.5319e+07 & 2.02203e+07 & 1.66408e+07 & 1.30266e+07 \\
$HC2$     & 183669200  & 204950271   & 211386386   & 230511930 \\}
\hline
\end{tabular}    
\end{center}
\end{table}

\section{Discussion and conclusion}

According to the average fitness found, Scuba Search outperforms the others local search heuristics on both $NKq$ and $TSPn$ as soon as neutrality is sufficient. However, it should be wondered whether efficiency of Scuba Search does have with the greatest number of evaluations. The number of evaluations for Scuba Search is lesser than the one for $HC2$. This last heuristic realizes a larger exploration of the neighborhood than $SS$: it pays attention to neighbors with same fitness and all the neighbors of the neighborhood too. However the average fitness found is worse than the one found by $SS$. So, considering the number of evaluations is not sufficient to explain good performance of $SS$. Whereas there is premature convergence towards local optima with $HC2$, $SS$ realizes a better compromise between exploration and exploitation by examining neutral neighbors. 

The main idea behind Scuba Search heuristic is to try to explicitly optimize evolvability on a neutral network before performing a qualitative step using a local search heuristic. If evolvability is almost constant on each neutral network, for instance as in the well-known Royal-Road landscape \cite{MIT-FOR-HOL:92}, $SS$ cannot perform neutral moves to increase evolvability and then have the same dynamic than $HC$. In this kind of problem, scuba search fails in all likehood.
\commentaire{Optimized evolvability needs evolvability not to be constant on a neutral network. For example in the well-known Royal-Road landscape, proposed by Mitchell et al.~\cite{MIT-FOR-HOL:92}, a high degree of neutrality leads evolvability to be constant on each neutral network.}

In order to reduce the evaluation cost of $SS$, one solution would be to choose a ``cheaper'' definition for evolvability: for example, the best fitness of $n$ neighbors randomly chosen or the first fitness of neighbor which improves the fitness of the current genotype. Antoher solution would be to change either the local search heuristic which evolvability or the one which allows to jump to a fitter solution. For instance, we could use Simulated Annealing or Tabu Search to optimize neutral network then jump to the first improvement met in the neighborhood.

This paper represents a first step demonstrating the potential interest in using the scuba search heuristic to optimize neutral landscape. Obviously we have to compare performances of this metaheuristic with other metaheuristics adapted to neutral landscape as Netcrawler \cite{barnett01netcrawling} or extrema selection \cite{STEW:01}. All these strategies use the neutrality in different ways to find good solution and may not have the same performances on all problems. $SS$ certainly works well when evolvability on neutral networks can be optimized.

\bibliography{Biblio}

\end{document}